\documentclass{article}
\usepackage{xcolor}
\definecolor{mydarkblue}{rgb}{0,0.08,0.45}
\usepackage{iclr2023_conference,times}
\usepackage{graphicx}
\usepackage{url}
\usepackage{tabularx}
\usepackage[toc,page,header]{appendix}
\usepackage{minitoc}
\usepackage{booktabs}
\usepackage{amsmath}
\usepackage{xcolor, soul}
\usepackage{multirow}
\usepackage{fontawesome}
\usepackage[colorlinks,citecolor=mydarkblue,urlcolor=mydarkblue,linkcolor=mydarkblue]{hyperref}
\usepackage{listings}
\usepackage{minted}

\title{\rule{\linewidth}{3pt}\\[1ex]  
Instruction-Following Evaluation in Function Calling for Large Language Models\\
\rule{\linewidth}{0.5pt}}             

\author{%
  Nikolai Skripko$^{1,2}$ \\
  \texttt{naskripko@edu.hse.ru} \\
  \\
  $^1$Higher School of Economics, Moscow \\
  $^2$SberDevices, Moscow
}

\begin{document}

\doparttoc
\faketableofcontents
\maketitle

\begin{abstract}
Function calling is a core capability of Large Language Models (LLMs), essential for AI agents. Existing benchmarks (e.g., BFCL \citep{patil2025bfcl}, $\tau^2$-Bench \citep{barres2025tau2}, ACEBench \citep{chen2025acebench}) evaluate argument correctness but do not test adherence to format instructions embedded in parameter descriptions, such as enclosing values in double quotes or using ISO date formats.

We introduce \textbf{IFEval-FC}, a benchmark inspired by IFEval \citep{zhou2023instructionfollowingevaluationlargelanguage}, which assesses precise instruction following in function calling. IFEval-FC encodes verifiable formats directly within JSON schema descriptions, such as “a value must not contain punctuation". It offers 750 test cases, each consisting of a function with an embedded format for one of its input parameters and a corresponding user query. The evaluation is fully algorithmic, ensuring objectivity, reproducibility, and scalability.

Our results indicate that even state-of-the-art proprietary models, such as GPT-5 \citep{openai2025gpt5} and Claude Opus 4.1 \citep{anthropic2025claude41}, frequently fail to adhere to basic formatting rules, highlighting a significant limitation for practical applications in real-world agent systems.
The complete codebase and data are publicly available at \faGithub\ \href{https://github.com/Skripkon/IFEval-FC}{\texttt{https://github.com/Skripkon/IFEval-FC}}
\end{abstract}

\section{Introduction}

Large Language Models (LLMs) are increasingly being deployed as core components of AI agents that interact with tools, APIs, and external systems via function calling. These agents leverage LLMs not only for natural language understanding and reasoning but also for structured API usage, tool execution, and decision-making in real-world domains.

A critical capability in this context is the model’s ability to correctly interpret and adhere to function signatures, particularly the formatting requirements specified in parameter descriptions within JSON schemas. For instance, a function schema may specify that a parameter \texttt{user\_name} must be “a string starting with a capital letter" or that a date must follow the ISO 8601 format. Despite their apparent simplicity, such format instructions are frequently overlooked or misinterpreted by LLMs, leading to invalid function calls and downstream failures in agent workflows.

Current benchmarks for function calling evaluate only functional correctness or API selection accuracy, often overlooking whether arguments are correctly formatted. This gap leaves a crucial dimension of agent robustness under-evaluated.

To address this, we introduce \textbf{IFEval-FC}, a benchmark inspired by IFEval that focuses on evaluating LLMs’ ability to follow \textit{verifiable format instructions} in the context of function calling. In IFEval-FC, each instruction is embedded directly into the description field of a parameter in a JSON schema (e.g., “should not include punctuation", “must be lowercase"). These instructions are designed to be objectively verifiable through algorithmic checks, enabling fully automated and reproducible evaluation without reliance on human or LLM-as-a-Judge \citep{gu2025surveyllmasajudge}.

\section{Verifiable instructions}
\label{sec:method}

The core innovation of IFEval-FC lies in its adaptation of ``verifiable instructions'' from the original IFEval benchmark. Unlike traditional function calling benchmarks that focus on functional correctness or API selection, IFEval-FC evaluates the model's ability to follow precise formatting constraints embedded within JSON schema parameter descriptions.

\subsection{Instruction Categories and Types}

We identified 19 distinct types of verifiable instructions, which we organized into seven major categories according to the nature of the constraints they impose (see Table 1).

A significant portion of our instruction types were adapted from the original IFEval benchmark, which focused on instruction following in general text generation tasks. Other types of instructions are introduced here for the first time (such as \texttt{Cyrillic Greek} or \texttt{Python List Format}).

\begin{table*}[t!]
\centering
\small
\begin{tabular}{l|p{2.9cm}|p{7.5cm}}
\toprule
\textbf{Instruction Group} & \textbf{Instruction} & \textbf{Description} \\
\midrule
Keywords & Keywords Presence & Requires inclusion / exclusion of specific keywords \\
\midrule
Keywords & Keyword Frequency & Specifies exact frequency requirements for keywords \\
\midrule
Keywords & Letter Frequency & Controls the frequency of specific letters \\
\midrule
Language & Cyrillic Greek & Restricts text to specific writing systems \\
\midrule
Length Constraints & Word Count & Controls the number of words in the response \\
\midrule
Length Constraints & Sentence Count & Manages sentence count requirements \\
\midrule
Detectable Content & Postscript & Requires specific postscript markers \\
\midrule
Detectable Content & Placeholder Count & Controls the number of placeholder markers \\
\midrule
Detectable Format & Spaces In Between & Enforces specific spacing patterns \\
\midrule
Detectable Format & Title Format & Requires specific formatting markers \\
\midrule
Detectable Format & Highlighted Sections Count & Controls markdown highlighting \\
\midrule
Detectable Format & Json Format & Requires valid JSON formatting \\
\midrule
Detectable Format & Python List Format & Enforces Python list syntax \\
\midrule
Case & All Uppercase & Requires all uppercase text \\
\midrule
Case & All Lowercase & Requires all lowercase text \\
\midrule
Case & N All Capital Words & Controls the number of all-caps words \\
\midrule
Start/End & End Phrase & Requires specific ending phrases \\
\midrule
Start/End & Quotation & Enforces quotation mark wrapping \\
\midrule
Punctuation & N Commas & Controls comma frequency \\
\bottomrule
\end{tabular}
\caption{Examples of verifiable format instructions adapted for function parameters. Each entry represents a category of instruction constraints encoded in a parameter description.}
\label{tab:instruction_checkers}
\end{table*}

\subsection{Dataset Creation}
\subsubsection{Functions}

A subset of our functions was sourced from the BFCL benchmark, providing real-world function schemas that we enhanced with our verifiable instruction constraints. Other functions were generated synthetically using GPT-5 through a carefully designed prompt engineering process. Our generation pipeline consisted of the following steps:

\begin{enumerate}
    \item \textbf{Domain Selection}: We curated 80 diverse domains representing real-world use cases where AI assistants might need to call functions.
    \item \textbf{Function Schema Generation}: For each domain, we used structured prompts to generate JSON schemas with specific requirements:
    \begin{itemize}
        \item Each function must include one \texttt{free-form parameter} for natural language input.
        \item The \texttt{free-form parameter} must be a string with no \texttt{enum} or \texttt{format} field.
    \end{itemize}
    This was done to simulate a situation similar to IFEval, where a model needs to generate natural language text when a specific format is imposed on it. Such a parameter allows the application of any format constraint, which is not always possible in BFCL, since not all formats are meaningful for specific arguments such as \texttt{user\_id}.
    
    \item \textbf{Instruction Injection}: We injected verifiable instructions into the description of one parameter of each function (\texttt{free-form parameter} for the generated functions and a randomly chosen string parameter with no \texttt{enum} or \texttt{format} field for the BFCL functions).
\end{enumerate}

\subsubsection{User Query Generation}

To ensure realistic evaluation scenarios, we generated five diverse user queries for each function using GPT-5. These queries were designed to satisfy the following criteria:
\begin{itemize}
    \item They are expressed in natural, conversational language.
    \item They contain values for all required parameters to call a function.
    \item They were generated based on the original functions (without embedded formats) so that users do not attempt to satisfy the format in advance; instead, it is the LLM’s task during function calling to transform a value into the required format.
    
\end{itemize}

Each function and its associated queries were further validated to ensure that the function was indeed necessary for completing the task. We verified this through a straightforward procedure: if an LLM failed to call the target function in response to all five queries, the task was deemed ill-posed or overly challenging and omitted. Conversely, tasks in which all five queries were answered correctly with ease were also omitted. An ensemble of LLMs was employed to increase the robustness of task selection.

We observed that Anthropic’s most recent models (e.g., Claude Opus 4.1) exhibited a significantly higher refusal rate when deciding whether to call a function compared to other models. Closer analysis revealed that these models often requested user clarification whenever subtle uncertainties arose (e.g., adherence to a specific format unknown to the user). To mitigate this issue and ensure fairness across evaluations, we added a system message explicitly instructing the model to always call a function:

\begin{lstlisting}[language=Python, caption={System message enforcing function invocation.}]
SYSTEM_MESSAGE = '''
YOU MUST CALL A FUNCTION NO MATTER WHAT.
NEVER ASK A USER TO SPECIFY OR CLARIFY ANYTHING.
ALWAYS CALL A FUNCTION.
'''.strip()
\end{lstlisting}

\subsection{Instruction Difficulty and Filtering}
During development, we discovered that certain instruction types (particularly trivial case constraints like all-uppercase/lowercase) were too easy for modern models, often achieving 90--100\% accuracy across models. To maintain the benchmark's discriminative power, we excluded these trivial instruction types from the final evaluation. The remaining instruction types represent a carefully curated set that provides meaningful differentiation between model capabilities.

\subsection{IFEval-FC metrics}
\label{section:ifeval_eval_metrics}
For a given response $r$ and a verifiable instruction $i$, we define the function that verifies whether the instruction is followed:

\begin{equation}
  \text{score}(r, i) = \begin{cases}
    \text{1}, & \text{if instruction is followed}.\\
    \text{0}, & \text{otherwise}.
  \end{cases}
\label{eq:is_followed}
\end{equation}

We evaluated several language models on IFEval-FC using Equation \ref{eq:is_followed}. Figure 1 presents the overall performance across all format constraint types. See Tables 2-4 in the appendix for a comprehensive breakdown by instruction.

\begin{figure}[htbp]
  \centering
  \includegraphics[width=0.8\linewidth]{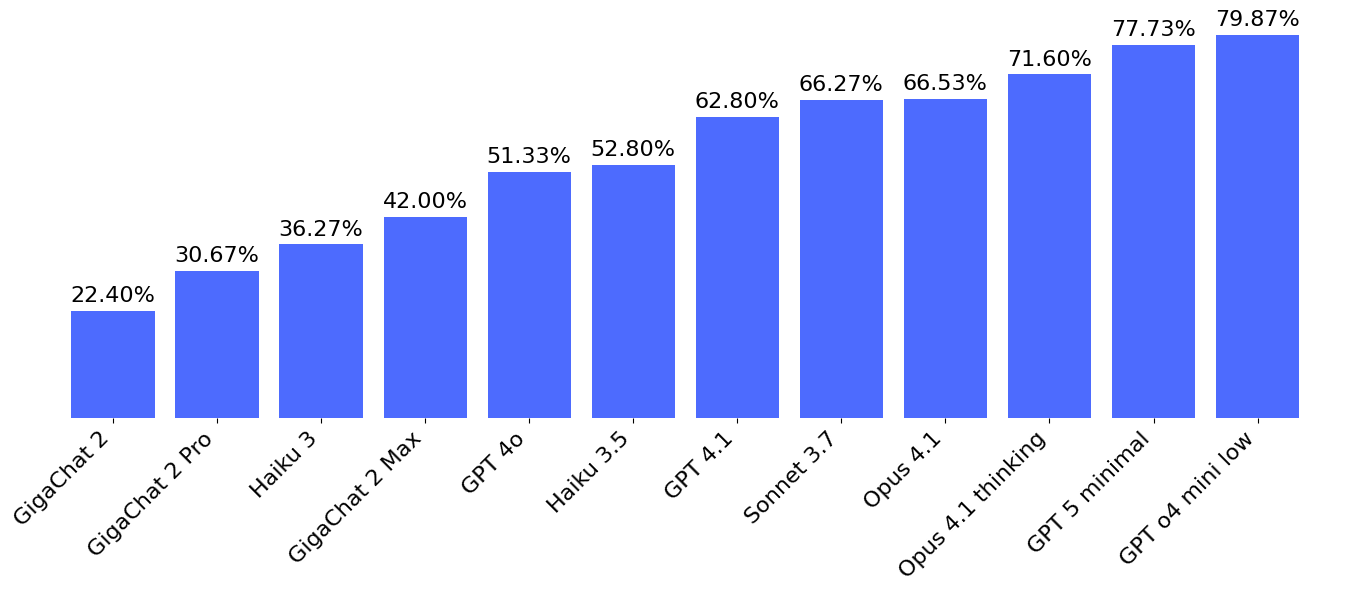}
  \caption{Overall performance of several language models on the IFEval-FC benchmark. Higher scores indicate better adherence to format constraints.}
  \label{fig:leaderboard}
\end{figure}

\section{Discussion and Future Work}
The results indicate that the latest models perform significantly better than their predecessors. However, no evaluated model surpassed 80\% accuracy, suggesting that precise instruction-following in function calling remains an open problem for LLMs, despite being a trivial task for humans.

To increase the benchmark’s difficulty, future work will expand the number of available functions and selecting more difficult samples from the originally generated ones. In the current setup, only one function is available for the model to call. A more challenging scenario would require the model to first select the correct function from a set of options, which may lead to a decrease in observed performance.

Additionally, while our current benchmark focuses on English-language function calls, future iterations could incorporate multilingual support, drawing inspiration from M-IFEval \citep{dussolle2025mifevalmultilingualinstructionfollowingevaluation} to assess cross-lingual instruction-following capabilities.

\clearpage

\bibliographystyle{iclr2023_conference}
\bibliography{main}

\clearpage
"
\part{Appendix}
\section{Detailed results}
\label{app:detailed_results}

\begin{table*}[h]
\centering
\small
\begin{tabular}{l|c|c|c}
\toprule
\textbf{Instruction} & \textbf{GigaChat 2} & \textbf{GigaChat 2 Pro} & \textbf{GigaChat 2 Max} \\
\midrule
Cyrillic Greek       & 22.00\% & 10.00\% & \textbf{\underline{50.00\%}} \\
Highlighted Sections Count & 38.00\% & 66.00\% & \textbf{\underline{72.00\%}} \\
Json Format          & 0.00\%  & 0.00\%  & 0.00\% \\
Keyword Frequency    & 28.00\% & 60.00\% & \textbf{\underline{64.00\%}} \\
Keywords Presence    & 54.00\% & 66.00\% & \textbf{\underline{84.00\%}} \\
Letter Frequency     & 12.00\% & 24.00\% & \textbf{\underline{42.00\%}} \\
N All Capital Words  & 30.00\% & 44.00\% & \textbf{\underline{46.00\%}} \\
N Commas             & 18.00\% & \textbf{\underline{40.00\%}} & 28.00\% \\
Placeholder Count    & 6.00\%  & 40.00\% & \textbf{\underline{58.00\%}} \\
Python List Format   & 10.00\% & 2.00\%  & \textbf{\underline{24.00\%}} \\
Quotation            & 26.00\% & 0.00\%  & \textbf{\underline{36.00\%}} \\
Sentence Count       & 28.00\% & 36.00\% & \textbf{\underline{58.00\%}} \\
Spaces In Between    & 2.00\%  & 4.00\%  & \textbf{\underline{8.00\%}} \\
Title Format         & 62.00\% & \textbf{\underline{64.00\%}} & 42.00\% \\
Word Count           & 0.00\%  & 4.00\%  & \textbf{\underline{18.00\%}} \\
\bottomrule
\end{tabular}
\caption{Evaluation results for GigaChat models.}
\label{tab:gigachat_results}
\end{table*}

\begin{table*}[h]
\centering
\small
\begin{tabular}{l|c|c|c|c|c}
\toprule
\textbf{Instruction} & \textbf{Haiku 3} & \textbf{Haiku 3.5} & \textbf{Sonnet 3.7} & \textbf{Opus 4.1} & \textbf{Opus 4.1 Thinking} \\
\midrule
Cyrillic Greek       & 30.00\% & 40.00\% & \textbf{\underline{44.00\%}} & 40.00\% & 34.00\% \\
Highlighted Sections Count & 48.00\% & 64.00\% & 86.00\% & 94.00\% & \textbf{\underline{100.00\%}} \\
Json Format          & 62.00\% & 30.00\% & 34.00\% & \textbf{\underline{68.00\%}} & \textbf{\underline{68.00\%}} \\
Keyword Frequency    & 36.00\% & 76.00\% & 88.00\% & \textbf{\underline{90.00\%}} & 86.00\% \\
Keywords Presence    & 50.00\% & 46.00\% & 86.00\% & 80.00\% & \textbf{\underline{90.00\%}} \\
Letter Frequency     & 22.00\% & 28.00\% & 38.00\% & 28.00\% & \textbf{\underline{54.00\%}} \\
N All Capital Words  & 28.00\% & \textbf{\underline{78.00\%}} & 76.00\% & 20.00\% & 14.00\% \\
N Commas             & 14.00\% & 12.00\% & 44.00\% & 52.00\% & \textbf{\underline{78.00\%}} \\
Placeholder Count    & 2.00\%  & 50.00\% & 76.00\% & 80.00\% & \textbf{\underline{92.00\%}} \\
Python List Format   & \textbf{\underline{94.00\%}} & 92.00\% & 68.00\% & 90.00\% & 90.00\% \\
Quotation            & 62.00\% & 34.00\% & 66.00\% & \textbf{\underline{70.00\%}} & 62.00\% \\
Sentence Count       & 26.00\% & 56.00\% & 72.00\% & 84.00\% & \textbf{\underline{86.00\%}} \\
Spaces In Between    & 10.00\% & 54.00 & \textbf{\underline{62.00\%}} & 8.00\%  & 24.00\% \\
Title Format         & 60.00\% & 78.00\% & 90.00\% & \textbf{\underline{100.00\%}} & \textbf{\underline{100.00\%}} \\
Word Count           & 0.00\%  & 54.00\% & 64.00\% & 94.00\% & \textbf{\underline{96.00\%}} \\
\bottomrule
\end{tabular}
\caption{Evaluation results for Anthropic models.}
\label{tab:anthropic_results}
\end{table*}

\begin{table*}[h]
\centering
\small
\begin{tabular}{l|c|c|c|c}
\toprule
\textbf{Instruction} & \textbf{GPT 4o} & \textbf{GPT 4.1} & \textbf{GPT 5 minimal} & \textbf{GPT o4 mini low} \\
\midrule
Cyrillic Greek       & 24.00\% & 36.00\% & 46.00\% & \textbf{\underline{70.00\%}} \\
Highlighted Sections Count & 58.00\% & 88.00\% & 86.00\% & \textbf{\underline{98.00\%}} \\
Json Format          & 40.00\% & 14.00\% & \textbf{\underline{58.00\%}} & 0.00\% \\
Keyword Frequency    & 80.00\% & 94.00\% & \textbf{\underline{98.00\%}} & 92.00\% \\
Keywords Presence    & 74.00\% & 90.00\% & 94.00\% & \textbf{\underline{98.00\%}} \\
Letter Frequency     & 28.00\% & 22.00\% & 36.00\% & \textbf{\underline{86.00\%}} \\
N All Capital Words  & 64.00\% & 84.00\% & 76.00\% & \textbf{\underline{90.00\%}} \\
N Commas             & 28.00\% & 42.00\% & 56.00\% & \textbf{\underline{82.00\%}} \\
Placeholder Count    & 12.00\% & 58.00\% & 84.00\% & \textbf{\underline{94.00\%}} \\
Python List Format   & 72.00\% & 94.00\% & \textbf{\underline{98.00\%}} & 84.00\% \\
Quotation            & 18.00\% & 46.00\% & \textbf{\underline{88.00\%}} & 56.00\% \\
Sentence Count       & 48.00\% & 60.00\% & \textbf{\underline{82.00\%}} & 78.00\% \\
Spaces In Between    & 76.00\% & 72.00\% & 88.00\% & \textbf{\underline{98.00\%}} \\
Title Format         & 76.00\% & 52.00\% & \textbf{\underline{94.00\%}} & \textbf{\underline{94.00\%}} \\
Word Count           & 72.00\% & \textbf{\underline{90.00\%}} & 82.00\% & 78.00\% \\
\bottomrule
\end{tabular}
\caption{Evaluation results for OpenAI models.}
\label{tab:openai_results}
\end{table*}

\label{fig:detailed_per_category_acc}

\end{document}